\documentclass[letterpaper]{article} 
\usepackage{aaai2026}  
\usepackage{times}  
\usepackage{helvet}  
\usepackage{courier}  
\usepackage[hyphens]{url}  
\usepackage{graphicx} 
\usepackage{natbib}  
\usepackage{caption} 
\usepackage{algorithm}
\usepackage{algorithmic}

\usepackage{multirow}
\usepackage{subfig}
\usepackage{amsfonts}
\usepackage{amsmath}

\usepackage{newfloat}
\usepackage{listings}
\DeclareCaptionStyle{ruled}{labelfont=normalfont,labelsep=colon,strut=off} 
\floatstyle{ruled}
\newfloat{listing}{tb}{lst}{}
\floatname{listing}{Listing}
\title{Parameter-Free Clustering via Self-Supervised Consensus Maximization (Extended Version)}
\author{
    Lijun Zhang\textsuperscript{\rm 1}\equalcontrib,
    Suyuan Liu\textsuperscript{\rm 1}\equalcontrib,
    Siwei Wang\textsuperscript{\rm 2},
    Shengju Yu\textsuperscript{\rm 1},
    Xueling Zhu\textsuperscript{\rm 3 \textdagger},\\
    Miaomiao Li\textsuperscript{\rm 4}\thanks{Corresponding authors.},
    Xinwang Liu\textsuperscript{\rm 1}
}
\affiliations{
    \textsuperscript{\rm 1}College of Computer Science and Technology, National University of Defense Technology, Changsha, 410073, China\\
    \textsuperscript{\rm 2}Academy of Military Sciences, Beijing, 100091, China\\
    \textsuperscript{\rm 3}Xiangya Hospital, Central South University, Changsha, 410008, China\\    
    \textsuperscript{\rm 4}College of Electronic Information and Electrical Engineering, Changsha University, Changsha, 410022, China\\
    \{junliz, suyuanliu, wangsiwei13\}@nudt.edu.cn, yu-shengju@foxmail.com, zhuxueling@csu.edu.cn, miaomiaolinudt@gmail.com, xinwangliu@nudt.edu.cn
}

\usepackage{bibentry}

\begin{document}

\maketitle

\begin{abstract}
Clustering is a fundamental task in unsupervised learning, but most existing methods heavily rely on hyperparameters such as the number of clusters or other sensitive settings, limiting their applicability in real-world scenarios. To address this long-standing challenge, we propose a novel and fully parameter-free clustering framework via \textbf{S}elf-supervised \textbf{C}onsensus \textbf{Max}imization, named SCMax. Our framework performs hierarchical agglomerative clustering and cluster evaluation in a single, integrated process. At each step of agglomeration, it creates a new, structure-aware data representation through a self-supervised learning task guided by the current clustering structure. We then introduce a nearest neighbor consensus score, which measures the agreement between the nearest neighbor-based merge decisions suggested by the original representation and the self-supervised one. The moment at which consensus maximization occurs can serve as a criterion for determining the optimal number of clusters. Extensive experiments on multiple datasets demonstrate that the proposed framework outperforms existing clustering approaches designed for scenarios with an unknown number of clusters.
\end{abstract}

\begin{links}
    \link{Code}{https://github.com/ljz441/2026-AAAI-SCMax}
    \link{Main Paper}{https://ojs.aaai.org/index.php/AAAI/article/view/40059}
\end{links}

\section{Introduction}

Clustering is a fundamental machine learning task that aims to partition unlabeled data into groups based on intrinsic similarities \cite{zhang2025max,zhou2025dpfmvc,liu2024sample,yu2024non,qin2024fast,wang2024evaluate}. As a typical unsupervised learning method, clustering has been widely applied in various domains such as image processing \cite{qiu2024multi,zhang2024another}, and text analysis \cite{he2025target,yang2025darec}, serving as a crucial tool for understanding data structures.

\begin{figure}[ht]
\begin{center}
{\includegraphics[width=0.9\columnwidth]{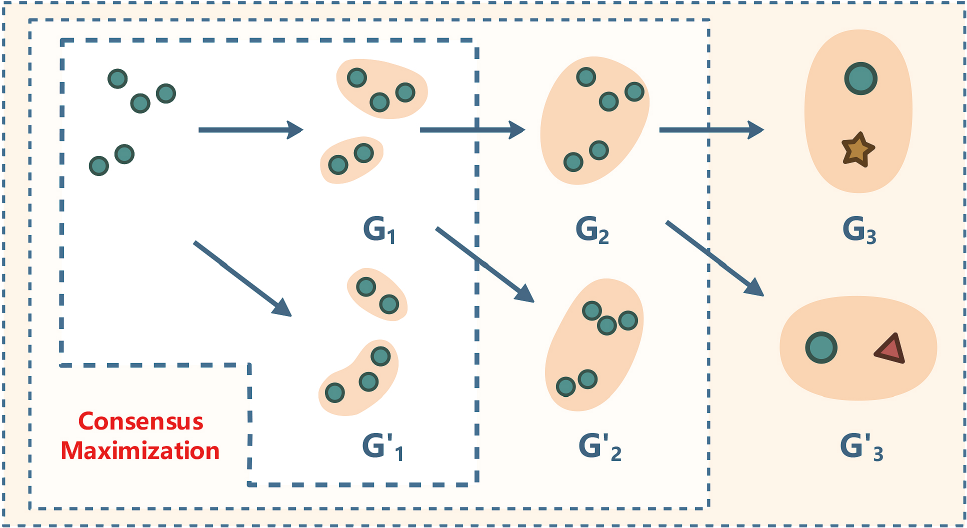}}
\caption{Motivation of Consensus Maximization. Here, $\mathbf{G}_i$ and $\mathbf{G'}_i$ denote the cluster structures from original and self-supervised representations, respectively. The optimal structure is determined when their consensus is maximized. For simplicity, each type of shape in $\mathbf{G}_3$ and $\mathbf{G'}_3$ is represented as a class set.}
\label{fig:fig1}
\end{center}
\end{figure}

However, existing clustering methods often heavily rely on hyperparameters such as the number of clusters or other sensitive settings, which poses significant challenges when dealing with tasks involving open and complex data structures. Unlike supervised learning, clustering lacks the guidance of class labels and, in real-world applications, it is generally infeasible to preset any priors. This reality has driven the emergence of parameter-free clustering techniques. In recent years, numerous clustering methods that do not rely on predefined cluster numbers or other sensitive settings have been proposed, including density-based \cite{abbas2021denmune,rodriguez2014clustering}, graph-based \cite{sun2024lsenet,peng2019comic,shah2017robust}, hierarchy-based \cite{yang2025autonomous,dang2021nearest,sarfraz2019efficient}, and probabilistic model-based \cite{wang2022dnb,yang2019vsb} approaches. Among them, hierarchy-based methods have become a prominent research direction due to their ability to naturally generate multi-level cluster structures during the clustering process, thereby producing multiple candidate values of $K$ with strong interpretability.

Although current parameter-free hierarchical clustering methods attempt to address the core issue of $K$-value selection, they do not entirely eliminate the reliance on priors. Many such methods, while not explicitly specifying $K$, introduce other sensitive parameter settings, such as distance thresholds or probability cutoffs, which still require manual tuning and can affect the stability and generalizability of the clustering results. Overall, these methods can typically be divided into two stages: cluster number generation and cluster structure evaluation. In the generation stage, common strategies often adopt split-and-merge techniques \cite{ronen2022deepdpm,liu2023reinforcement,dai2024multi}. However, these approaches usually imply assumptions about the initial number of clusters and can only explore a limited set of $K$ values in practice, making it difficult to cover all potential clustering structures. In the evaluation stage, common techniques such as the elbow method \cite{shi2021quantitative,schubert2023stop} often rely on smoothed or ambiguous evaluation curves that require manually defined thresholds or subjective identification of "elbow points," which can be unstable in high-dimensional or complex data scenarios. Therefore, how to automatically generate and evaluate cluster structures without any prior knowledge or manual settings remains a key challenge in achieving truly parameter-free clustering.

To address the above issues, we propose a fully parameter-free clustering framework via \textbf{S}elf-supervised \textbf{C}onsensus \textbf{Max}imization, named SCMax. Falling under the category of hierarchical clustering, SCMax achieves the complete process from cluster number generation to cluster structure evaluation without relying on any prior settings. Specifically, in the generation stage, SCMax constructs nearest neighbor graphs to guide the cluster merging process and automatically produces a set of candidate cluster structures. Unlike existing methods, SCMax does not assume an initial number of clusters or restrict the search space, enabling the direct and efficient generation of multiple candidates and significantly reducing the $K$-value search space. Furthermore, based on the principle of nearest neighbor stability, we design a Nearest Neighbor Consensus (NNC) evaluation metric to measure the agreement of merging decisions between the original and self-supervised feature representations. The optimal number of clusters is automatically determined at the point of consensus maximization. Compared to traditional methods, this metric avoids reliance on subjective parameters such as thresholds or elbow points. The entire process is fully automated, achieves strong generalization and robustness, and embodies the essence of “parameter-free” clustering. The motivation is illustrated in Fig.~\ref{fig:fig1}. The main contributions of this work are summarized as follows.

\begin{itemize}
    \item We propose a fully parameter-free clustering framework that performs hierarchical agglomerative clustering and cluster evaluation in a single integrated process. This framework does not require any hyperparameters of the number of clusters or other sensitive settings.
    \item We design a cluster evaluation metric based on nearest neighbor consensus, which measures the agreement between the nearest neighbor-based merge decisions derived from the original and self-supervised representations. The moment at which consensus maximization occurs can serve as a criterion for determining the optimal number of clusters.
    \item Extensive experiments on multiple datasets demonstrate that the proposed framework outperforms existing clustering approaches designed for scenarios with an unknown number of clusters.
\end{itemize}

\section{Related Work}

\begin{figure*}[ht]
\begin{center}
{\includegraphics[width=0.85\textwidth]{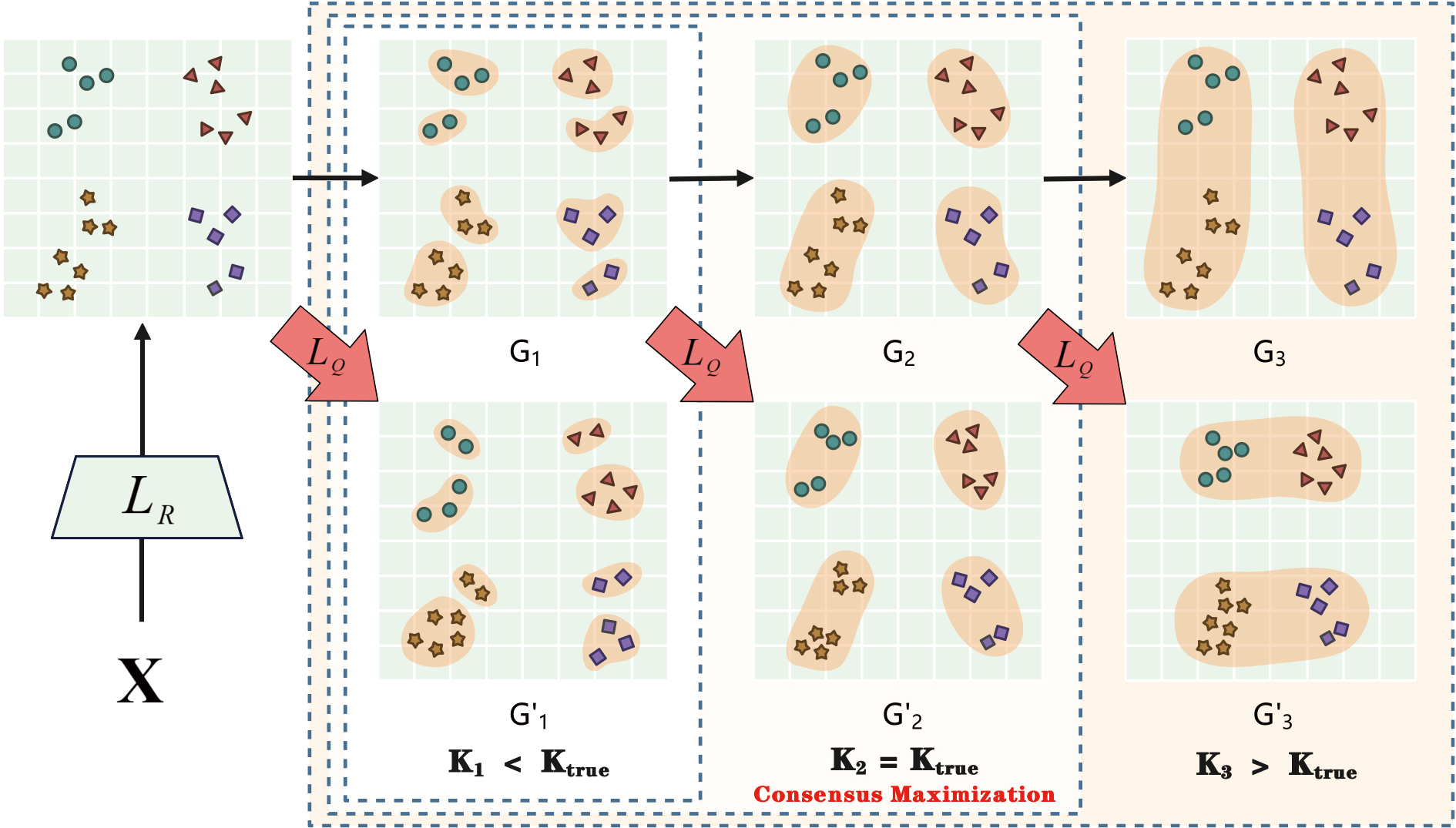}}
\caption{The proposed SCMax framework.}
\label{fig:fig2}
\end{center}
\end{figure*}

This section focuses on the branch of hierarchical clustering under the umbrella of parameter-free clustering, with particular emphasis on two key aspects: cluster number generation and cluster structure evaluation.

\subsection{Cluster Number Generation}

Within the framework of parameter-free clustering, traditional hierarchical clustering methods mainly adopt two types of cluster number generation strategies: bottom-up and top-down. The former, such as agglomerative clustering \cite{li2024multi,xing2021learning}, starts with a larger number of clusters and progressively merges the most similar ones; the latter, such as divisive clustering \cite{bagirov2023finding,liu2023reinforcement,dai2024multi}, begins with fewer clusters and recursively splits them into smaller sub-clusters. In addition, some methods combine the strengths of both approaches, such as split-and-merge strategy \cite{zhao2024parameter,xiao2023deep,ronen2022deepdpm}, which dynamically decides whether to split or merge clusters based on an initial assumption of the number of clusters. While these methods are capable of constructing a hierarchical structure of clustering results, they still face several practical challenges. On one hand, many of them heavily rely on the initial number of clusters. If this initial value deviates significantly from the true number of clusters, it may lead to inefficient clustering or misguide the subsequent structural exploration, resulting in poor candidate cluster structures. On the other hand, most approaches require a predefined search space for the number of clusters in order to avoid the high computational cost of brute-force enumeration, which inevitably limits their flexibility and generalization ability.

To address these challenges, Nearest Neighbor Clustering (NN clustering) methods have attracted increasing attention in recent years \cite{yang2025autonomous,dang2021nearest,sarfraz2019efficient}. Compared to traditional strategies, NN clustering offers superior computational efficiency, with complexity reduced to $O(N \log N)$. Without the need to assume an initial number of clusters or constrain the search space, NN clustering can directly and efficiently generate multiple candidate cluster structures, significantly reducing the $K$-value search space. These methods not only improve runtime performance and clustering quality but also scale well to large datasets, demonstrating broad applicability in hierarchical clustering tasks.

\subsection{Cluster Structure Evaluation}

In hierarchical clustering frameworks, the core challenge of parameter-free clustering lies in how to automatically identify the optimal partition from a set of hierarchical candidate cluster structures, which is crucial for ensuring clustering performance. Traditional approaches such as the Elbow Method \cite{article,shi2021quantitative,schubert2023stop} typically rely on manually set thresholds or visually identified inflection points to determine when to stop the clustering process. However, such methods are often highly sensitive to subjective settings, lack adaptive capabilities, and are vulnerable to noise, which can lead to unstable and unreliable evaluation results.

To overcome these limitations, we observe a phenomenon referred to as nearest-neighbor stability in the actual feature representation space and propose a nearest-neighbor consensus-based cluster structure evaluation metric. This metric measures the consistency of nearest-neighbor merging decisions between the original features and those obtained via self-supervised learning. The moment when this consensus maximization occurs corresponds to the optimal cluster structure. This metric is entirely free of manually set hyperparameters and can adaptively capture the intrinsic structure of the data, enabling truly parameter-free, stable, and reliable cluster structure evaluation.

\section{Methodology}

In this section, we propose a fully parameter-free clustering framework termed SCMax. The overall framework is illustrated in Fig.~\ref{fig:fig2}.

\subsection{Autoencoder Module}

Given a dataset $\mathbf{X} \in \mathbb{R}^{N \times D}$, where $N$ denotes the number of samples and $D$ denotes the feature dimension. For the original features $\mathbf{X}$, there may exist redundancy and noise. Therefore, we use an autoencoder to nonlinearly map $\mathbf{X}$ into a customizable feature space to extract more representative feature embeddings. Specifically, we denote the encoder and decoder as $E(\mathbf{X}, \theta)$ and $D(\mathbf{X}, \phi)$, where $\theta$ and $\phi$ are network parameters. Based on this, the mapped $L$-dimensional latent feature representation is $\mathbf{Z} = E(\mathbf{X}) \in \mathbb{R}^{N \times L}$, and the reconstructed representation is $\hat{\mathbf{X}} = D(E(\mathbf{X}))$. The reconstruction loss between input $\mathbf{X}$ and output $\hat{\mathbf{X}}$ is denoted as $L_R$. Thus, the autoencoder’s loss is formulated as:
\begin{equation}
L_{R} = \sum_{i=1}^{N} \| \mathbf{X}_{i} - \hat{\mathbf{X}}_{i} \|_{2}^{2} \text{ .} \tag{1}
\end{equation}

\subsection{Cluster Number Generation Module}

Based on the dimensionally reduced feature representation $\mathbf{Z}$, we begin cluster number generation. In SCMax, we adopt nearest neighbor merging to generate cluster candidates due to its efficiency, as it aggregates data using only local neighbor relations without constructing a global distance matrix. The resulting partitions have been shown to closely match true clusters and perform well across various tasks \cite{sarfraz2019efficient,dang2021nearest,yang2025autonomous}. Specifically, each sample is initially treated as an individual cluster, forming the first-level cluster structure. We then recursively merge these clusters, where at each step, each cluster performs merging operations guided by the nearest neighbor graph, thereby generating a new cluster structure. To efficiently construct the nearest neighbor graph, we employ an approximate nearest neighbor search method (i.e., KD-Tree). Each cluster is represented by the mean of its sample vectors, and the nearest class neighbors are computed based on these mean vectors. Unlike sample-level nearest neighbor merging, this approach performs merging at the class level, which simplifies computation and maintains a complexity of $O(K \log K)$, where $K$ denotes the number of clusters. We denote the nearest neighbor graph as $\mathbf{A} \in \mathbb{R}^{K \times 1}$. Given the integer indices $\mathbf{A}$ representing the nearest neighbor of each class, we can construct a symmetric adjacency matrix $\mathbf{G} \in \mathbb{R}^{K \times K}$ in linear time as follows:
\begin{equation}
\mathbf{G}(i,j) = 
\begin{cases} 
1, & \text{if } j = \mathbf{A}_i\text{ or } \mathbf{A}_j = i \text{ or } \mathbf{A}_i = \mathbf{A}_j \\ 
0, & \text{otherwise}
\end{cases} \text{ ,} \tag{2}
\end{equation}

\noindent where $\mathbf{A}_i$ and $\mathbf{A}_j$ denote the nearest neighbors of classes $i$ and $j$, respectively. The adjacency matrix $\mathbf{G}$ captures the inter-class relationships, and its connected components correspond to the resulting clusters. Since $\mathbf{G}$ describes class-level relations, a reverse label mapping is required to propagate the clustering results back to the sample level, thereby obtaining the final cluster label vector $\mathbf{Q} \in \mathbb{R}^{N \times 1}$. This clustering formulation is extremely simple and parameter-free, directly producing a set of candidate cluster structures from the data.

\subsection{Cluster Structure Evaluation Module}

After obtaining multiple hierarchical candidate cluster structures $\mathbf{G}$, the core problem of the evaluation module is how to select the optimal clustering structure from these candidates. Before formally introducing the cluster structure evaluation module in SCMax, we first need to introduce the core concept of “nearest neighbor stability.” Specifically, when perturbations are applied to the representation $\mathbf{Z}$, the adjacency relationships of classes in the nearest neighbor graph may change. We define the degree of change in nearest neighbor graph structure before and after perturbation as the classes’ nearest neighbor stability, which indicates the robustness of the local structure to perturbations.

Next, we provide the definition and implementation of the perturbation. In SCMax, perturbations are not based on random noise but are implemented by introducing label $\mathbf{Q}$-based contrastive constraints. These constraints impose structural interventions on the fixed feature representation $\mathbf{Z}$, indirectly altering the adjacency relationships among classes to simulate perturbations. Under the contrastive learning constraint, we aim to pull together samples of the same class while pushing apart samples from different classes. Hence, we define the following structural contrastive loss:
\begin{equation}
L_{Q} = \frac{1}{|\mathcal{P}|} \sum_{(i,j) \in \mathcal{P}} \mathbf{D}_{ij} - \frac{1}{|\mathcal{N}|} \sum_{(i,j) \in \mathcal{N}} \mathbf{D}_{ij} \text{ ,} \tag{3}
\end{equation}

\noindent where $\mathcal{P} = \{(i, j) \mid q_i = q_j,\ i \ne j\}$ denote the set of positive (same-class) sample pairs, $\mathcal{N} = \{(i, j) \mid q_i \ne q_j\}$ denote the set of negative (different-class) sample pairs, and $\mathbf{D}_{ij}$ represents the Euclidean distance between $z_i$ and $z_j$, where $\mathbf{D} \in \mathbb{R}^{B \times B}$ and $B$ is the batch size. To avoid $O(N^2)$ complexity in computation, SCMax does not perform full-sample contrastive learning but adopts a mini-batch training strategy. That is, at each training step, a local Euclidean distance matrix is constructed within the current batch, significantly reducing computation and memory costs. This loss function uses label $\mathbf{Q}$ as supervision to guide the fixed representation $\mathbf{Z}$ to undergo structured changes in semantic space, achieving the perturbation goal. Unlike random noise perturbations, this method’s intervention in local structure is directional and controllable, effectively influencing nearest neighbor stability and providing a clean and explicit baseline for subsequent cluster structure evaluation.

At this point, after perturbing the fixed representation $\mathbf{Z}$ under the guidance of cluster labels $\mathbf{Q}$, we obtain a set of new self-supervised representations $\mathbf{Z'} = \{ {\mathbf{Z'}_1, \mathbf{Z'}_2, ..., \mathbf{Z'}_i}\}$, where each $\mathbf{Z'}_i$ corresponds to the perturbation guided by $\mathbf{Q}_i$. Based on the fixed representation $\mathbf{Z}$, we find the nearest neighbors within each class in $\mathbf{G}_i$ and perform merging to get the next cluster structure $\mathbf{G}_{i+1}$. Similarly, based on the self-supervised representation $\mathbf{Z'}_i$, we find the nearest neighbors in $\mathbf{G}_i$ and merge to obtain a new cluster structure $\mathbf{G'}_{i+1}$. So far, we have two sets of cluster structures: $\mathbf{G} = \{\mathbf{G}_1, \mathbf{G}_2, ..., \mathbf{G}_i\}$ from the fixed representation $\mathbf{Z}$, and $\mathbf{G'} = \{\mathbf{G'}_2, ..., \mathbf{G'}_i\}$ from the self-supervised representation $\mathbf{Z'}$. By measuring the consistency between the merging decisions of nearest neighbors executed on the original representation $\mathbf{Z}$ and the self-supervised representation $\mathbf{Z'}$—that is, the similarity between cluster structures $\mathbf{G}_i$ and $\mathbf{G'}_i$—we define the Nearest Neighbor Consensus (NNC) metric as follows:
\begin{equation}
\mathrm{NNC}(\mathbf{G}_i, \mathbf{G'}_i) = \max_{\pi \in \mathcal{S}} \frac{1}{N} \sum_{j=1}^N \mathbf{1}\left(\mathbf{Q}_i(j) = \pi(\mathbf{Q'}_i(j))\right) \text{ ,} \tag{4}
\end{equation}

\noindent where $\mathbf{Q}_i$ and $\mathbf{Q'}_i$ denote the cluster assignment labels for each sample in the clusters $\mathbf{G}_i$ and $\mathbf{G'}_i$, respectively; the indicator function $\mathbf{1}(\cdot)$ returns 1 if the condition is true and 0 otherwise; $\pi$ represents the set $\mathcal{S}$ of all possible label mappings under the Hungarian algorithm for label alignment.

\begin{algorithm}[tb]
\caption{SCMax}
\label{alg:algorithm}
\begin{algorithmic}[1] 
\STATE \textbf{Input}: Dataset $\mathbf{X} \in \mathbb{R}^{N \times D}$
\STATE \textbf{Output}: Cluster number $K^*$, cluster labels $\mathbf{Q^*}$
\STATE Compute latent representation $\mathbf{Z} \in \mathbb{R}^{N \times L}$ by Eq.(1)
\STATE Initialize each sample as a singleton cluster
\STATE Get $\mathbf{G}_1$ via nearest neighbor merging on $\mathbf{Z}$ by Eq.(2)
\STATE Derive $K_1$, $\mathbf{Q}_1$ from connected components in $\mathbf{G}_1$
\STATE Let $i = 1$
\WHILE{at least three clusters exist in $\mathbf{G}_i$}
    \STATE Update $\mathbf{G}_{i+1}$ via nearest neighbor merging based on $\mathbf{Z}$ by Eq.(2)
    \STATE Obtain $K_{i+1}$ and $\mathbf{Q}_{i+1}$ by computing connected components in $\mathbf{G}_{i+1}$
    \STATE Generate perturbed representation $\mathbf{Z'}$ based on $\mathbf{Q}_i$ by Eq.(3)
    \STATE Compute $\mathbf{G'}_{i+1}$ via nearest neighbor merging based on $\mathbf{Z'}$ by Eq.(2)
    \STATE Compute NNC score between $\mathbf{G}_{i+1}$ and $\mathbf{G'}_{i+1}$ by Eq.(4)
    \STATE Record $K_{i+1}$, $\mathbf{Q}_{i+1}$ and corresponding NNC score 
    \STATE $i \leftarrow i + 1$
\ENDWHILE
\STATE Select $K^*$, $\mathbf{Q^*}$ with the highest NNC score
\STATE \textbf{return} $K^*$, $\mathbf{Q^*}$
\end{algorithmic}
\end{algorithm}

Finally, we provide an in-depth discussion and analysis of the NNC metric, explaining why the moment when consensus reaches its maximum corresponds to the optimal clustering structure. The detailed process is illustrated in the Fig.~\ref{fig:fig2}.

\begin{itemize}
    \item \textbf{When the cluster number $K_i$ corresponding to $\mathbf{G}_i$ is greater than the true cluster number}: perturbations applied on the fixed representation $Z$ at the previous step significantly affect the nearest neighbor relationships among intra-class samples of the true cluster structure. Since samples in true clusters are close to each other, even slight perturbations can break the original adjacency relations, indicating unstable nearest neighbor relations at $K_{i-1}$. Consequently, the nearest neighbor graph constructed from the perturbed self-supervised representation $\mathbf{Z'}$ significantly differs from that constructed from $\mathbf{Z}$, resulting in low similarity between $\mathbf{G}_i$ and $\mathbf{G'}_i$.
    \item \textbf{When $K_i$ equals the true cluster number}: perturbations applied on $\mathbf{Z}$ at the previous step do not significantly affect intra-class adjacency. At this point, $K_{i-1}$ approaches the true cluster number, with most intra-class samples merged into the same cluster, greatly reducing the number of intra-class nearest neighbors and leading to a stable adjacency relationships. Therefore, the difference between perturbed and original nearest neighbor graphs is minimal, and the similarity between $\mathbf{G}_i$ and $\mathbf{G'}_i$ reaches its maximum.
    \item \textbf{When $K_i$ is less than the true cluster number}: perturbations applied on $\mathbf{Z}$ disrupt the inter-class boundary structure of the true cluster. Since true clusters usually have certain gaps, perturbations blur or overlap these gaps, causing samples from different true clusters to be incorrectly grouped together. This breaks the original inter-class boundaries, making nearest neighbor relations unstable again at $K_{i-1}$. As a result, the difference between perturbed and original nearest neighbor graphs becomes significant again, and the similarity between $\mathbf{G}_i$ and $\mathbf{G'}_i$ decreases.
\end{itemize}

In summary, when the nearest neighbor consensus reaches its maximum, the corresponding cluster structure $\mathbf{G^*}$ precisely reflects the optimal cluster distribution of the data, and the corresponding cluster number $K^*$ and label set $\mathbf{Q^*}$ can be regarded as the final clustering result. It can be seen that the proposed NNC metric does not depend on any prior hyperparameters and can adaptively evaluate the intrinsic structure of the data.

\subsection{Computational Complexity Analysis}

The pseudo-code of SCMax is presented in Algorithm \ref{alg:algorithm}. This subsection focuses on analyzing the main computational and memory overheads. Regarding time complexity, constructing the class-level nearest neighbor index using a KD-Tree for $K$ classes requires $O(K \log K)$ time. The symmetric adjacency matrix $\mathbf{G}$ is then constructed in linear time, $O(K)$. To assess local consistency, SCMax computes the Euclidean distance matrix within each mini-batch of size $B$, which incurs a cost of $O(B^2)$. During the computation of the NNC score, the Hungarian matching algorithm introduces an additional complexity of $O(K^3)$. Therefore, the total time complexity is $O(K \log K + K + B^2 + K^3)$. As for space complexity, the nearest neighbor index requires $O(K)$ space, the adjacency matrix occupies $O(K^2)$, and the distance matrix within each mini-batch takes $O(B^2)$. In addition, constructing the confusion matrix for computing the NNC score requires $O(K^2)$ space. Consequently, the overall space complexity is $O(K + 2K^2 + B^2)$.

\begin{table}[t]
\centering
\setlength{\tabcolsep}{1mm}
\begin{tabular}{cccc}
\hline
Dataset & Samples & Dimension & Cluster \\
\hline
MSRCv1 & 210 & 1302 & 7 \\
HW2 & 2000 & 784 & 10 \\
Wiki & 2866 & 10 & 10 \\
MNIST & 5000 & 784 & 10 \\
Cifar10 & 50000 & 1024 & 10 \\
Cifar100 & 50000 & 512 & 100 \\
Fashion & 60000 & 1280 & 10 \\
YTF20 & 63896 & 512 & 20 \\
\hline
\end{tabular}
\caption{Datasets description.}
\label{tab:dataset_description}
\end{table}

\begin{table*}[!ht]
\centering
\setlength{\tabcolsep}{1mm}
\begin{tabular}{c|ccccccccc|c}
\hline
\multirow{2}{*}{Datasets} & FINCH & COMIC & BP & DenMune & DeepDPM & MPAASL & TC & Gauging-$\delta$ & AFCL & \multirow{2}{*}{SCMax} \\ 
                          & CVPR'19 & ICML'19 & TPAMI'20 & PR'21 & CVPR'22 & ICML'24 & TPAMI'25 & TPAMI'25 & AAAI'25 &                        \\ \hline
\multicolumn{11}{c}{ACC}                                                                       \\ \hline
MSRCv1      & 0.2524  & 0.1667  & 0.2762   & 0.4048  & 0.1429  & 0.4476  & 0.3048  & 0.2333  & 0.0310  & \textbf{0.6238}  \\
HW2         & 0.1975  & 0.1045  & 0.1410   & 0.2005  & 0.1000  & 0.5980  & 0.4725  & 0.1000  & 0.0005  & \textbf{0.6115}  \\
Wiki        & 0.5286  & 0.1207  & 0.4833   & 0.1019  & 0.1574  & 0.5056  & 0.4682  & 0.1574  & 0.0052  & \textbf{0.5967}  \\
MNIST & 0.1968  & 0.1028  & 0.5004   & 0.0910  & 0.1000  & 0.5946  & 0.4752  & 0.1000  & 0.1000  & \textbf{0.6004}  \\
Cifar10    & 0.7117  & OOM      & OOM       & 0.0193  & 0.8178  & OOM      & OOM  & OOM & OOM     & \textbf{0.8703}  \\
Cifar100   & 0.5999  & OOM      & OOM       & 0.0742  & 0.3100  & OOM      & OOM  & OOM & OOM     & \textbf{0.9061}  \\
Fashion      & 0.2933  & OOM      & OOM       & 0.0183  & 0.3652  & OOM      & OOM  & OOM & OOM     & \textbf{0.7542}  \\
YTF20       & 0.1238  & OOM      & OOM       & 0.0253  & 0.4890  & OOM      & OOM   & OOM & OOM    & \textbf{0.6493}  \\ \hline
\multicolumn{11}{c}{NMI}                                                                       \\ \hline
MSRCv1      & 0.2185  & 0.0454  & 0.1335   & 0.5479  & 0.0000  & \textbf{0.5888}  & 0.5567  &0.0986  & 0.2936  & 0.5597  \\
HW2         & 0.3480  & 0.0128  & 0.0731   & 0.5616  & 0.0000  & \textbf{0.7459}  & 0.5998  &0.0000  & 0.2198  & 0.6436  \\
Wiki        & 0.5448  & 0.4654  & 0.5209   & 0.4151  & 0.0000  & 0.5322  & 0.5284  & 0.0000  & 0.0061  & \textbf{0.5609}  \\
MNIST & 0.3562  & 0.0109  & 0.6005   & 0.4946  & 0.0000  & \textbf{0.7385}  & 0.6321  & 0.0000  & 0.0000 & 0.7021  \\
Cifar10    & 0.7262  & OOM      & OOM       & 0.3707  & 0.7474  & OOM      & OOM   & OOM & OOM    & \textbf{0.7714}  \\
Cifar100   & 0.8246  & OOM      & OOM       & 0.6712  & 0.8368  & OOM      & OOM   & OOM & OOM    & \textbf{0.9801}  \\
Fashion      & 0.4830  & OOM      & OOM       & 0.3587  & 0.6511  & OOM      & OOM  & OOM & OOM     & \textbf{0.7479}  \\
YTF20       & 0.0503  & OOM      & OOM       & 0.4490  & \textbf{0.8126}  & OOM      & OOM    & OOM & OOM   & 0.8065  \\ \hline
\multicolumn{11}{c}{PUR}                                                                       \\ \hline
MSRCv1      & 0.2524  & 0.1667  & 0.2810   & 0.6762  & 0.1429  & 0.7762  & \textbf{0.7905}  & 0.2333  & 0.5262  & 0.6476  \\
HW2         & 0.1975  & 0.1065  & 0.1410   & \textbf{0.8625}  & 0.1000  & 0.6960  & 0.7785  & 0.1000  & 0.2295  & 0.6115  \\
Wiki        & 0.6350  & \textbf{0.9337}  & 0.6308   & 0.6933  & 0.1574  & 0.6497  & 0.6643  & 0.1574  & 0.2102  & 0.6403  \\
MNIST & 0.1968  & 0.1052  & 0.5004   & 0.8708  & 0.1000  & \textbf{0.8738}  & 0.8228  & 0.1000  & 0.2178  & 0.7658  \\
Cifar10    & 0.7117  & OOM      & OOM       & 0.8361  & 0.8376  & OOM      & OOM  & OOM & OOM     & \textbf{0.8922}  \\
Cifar100   & 0.5999  & OOM      & OOM       & 0.9055  & 0.3100  & OOM      & OOM   & OOM & OOM    & \textbf{0.9318}  \\
Fashion      & 0.2933  & OOM      & OOM       & 0.8161  & 0.8358  & OOM      & OOM  & OOM & OOM     & \textbf{0.8497}  \\
YTF20       & 0.1238  & OOM      & OOM       & 0.8345  & \textbf{0.9711}  & OOM      & OOM   & OOM & OOM    & 0.8071  \\ \hline
\multicolumn{11}{c}{F-score}                                                                   \\ \hline
MSRCv1      & 0.1593  & 0.0803  & 0.2029   & 0.2045  & 0.0357  & 0.2066  & 0.0797  & 0.1322  & 0.4419  & \textbf{0.5698}  \\
HW2         & 0.0738  & 0.0247  & 0.0799   & 0.0234  & 0.0182  & 0.5175  & 0.2430  & 0.0182  & 0.2695 & \textbf{0.5538}  \\
Wiki        & 0.3957  & 0.0011  & 0.3574   & 0.0069  & 0.0272  & 0.2895  & 0.2534  & 0.0272  & 0.1822 & \textbf{0.5538}  \\
MNIST & 0.0741  & 0.0121  & 0.4255   & 0.0035  & 0.0182  & 0.3603  & 0.2318  & 0.0182 & 0.1540  & \textbf{0.4553}  \\
Cifar10    & 0.6484  & OOM      & OOM       & 0.0001  & 0.7933  & OOM      & OOM   & OOM & OOM    & \textbf{0.8001}  \\
Cifar100   & 0.5778  & OOM      & OOM       & 0.0029  & 0.1691  & OOM      & OOM    & OOM & OOM   & \textbf{0.8476}  \\
Fashion      & 0.1854  & OOM      & OOM       & 0.0000  & 0.1550  & OOM      & OOM   & OOM & OOM    & \textbf{0.6187}  \\
YTF20       & 0.0332  & OOM      & OOM       & 0.0001  & 0.1543  & OOM      & OOM   & OOM & OOM    & \textbf{0.5671}  \\ \hline
\end{tabular}
\caption{Non-$K$ methods clustering performance comparison results, where the best results are highlighted in bold and OOM denotes out of memory error.}
\label{tab:clustering_performance}
\end{table*}

\begin{table*}[!ht]
\centering
\setlength{\tabcolsep}{1mm}
\begin{tabular}{ccccccccc}
\hline
\multicolumn{1}{c|}{Methods}        & MSRCv1          & HW2             & Wiki            & MNIST           & Cifar10         & Cifar100        & Fashion         & YTF20           \\ \hline
\multicolumn{9}{c}{ACC}                                                                                                                                                             \\ \hline
\multicolumn{1}{c|}{DCGL (AAAI'24)} & \textbf{0.6810} & \textbf{0.6185} & 0.5534          & \textbf{0.6246} & OOM             & OOM             & OOM             & OOM             \\
\multicolumn{1}{c|}{DMAC (AAAI'25)} & 0.4905          & 0.5285          & 0.1574          & 0.1000          & OOM             & OOM             & OOM             & OOM             \\
\multicolumn{1}{c|}{SCMax}          & 0.6238          & 0.6115          & \textbf{0.5967} & 0.6004          & \textbf{0.8703} & \textbf{0.9061} & \textbf{0.7542} & \textbf{0.6493} \\ \hline
\multicolumn{9}{c}{NMI}                                                                                                                                                             \\ \hline
\multicolumn{1}{c|}{DCGL (AAAI'24)} & \textbf{0.6143} & \textbf{0.6804} & 0.5356          & 0.6621          & OOM             & OOM             & OOM             & OOM             \\
\multicolumn{1}{c|}{DMAC (AAAI'25)} & 0.4841          & 0.4708          & 0.0000          & 0.0000          & OOM             & OOM             & OOM             & OOM             \\
\multicolumn{1}{c|}{SCMax}          & 0.5597          & 0.6436          & \textbf{0.5609} & \textbf{0.7021} & \textbf{0.7714} & \textbf{0.9801} & \textbf{0.7479} & \textbf{0.8065} \\ \hline
\multicolumn{9}{c}{PUR}                                                                                                                                                             \\ \hline
\multicolumn{1}{c|}{DCGL (AAAI'24)} & \textbf{0.7000} & \textbf{0.6940} & 0.6179          & 0.6712          & OOM             & OOM             & OOM             & OOM             \\
\multicolumn{1}{c|}{DMAC (AAAI'25)} & 0.5381          & 0.5425          & 0.1574          & 0.1000          & OOM             & OOM             & OOM             & OOM             \\
\multicolumn{1}{c|}{SCMax}          & 0.6476          & 0.6115          & \textbf{0.6403} & \textbf{0.7658} & \textbf{0.8922} & \textbf{0.9318} & \textbf{0.8497} & \textbf{0.8071} \\ \hline
\multicolumn{9}{c}{F-score}                                                                                                                                                         \\ \hline
\multicolumn{1}{c|}{DCGL (AAAI'24)} & \textbf{0.6771} & 0.5322          & 0.5428          & \textbf{0.6206} & OOM             & OOM             & OOM             & OOM             \\
\multicolumn{1}{c|}{DMAC (AAAI'25)} & 0.3776          & 0.5349          & 0.0272          & 0.0182          & OOM             & OOM             & OOM             & OOM             \\
\multicolumn{1}{c|}{SCMax}          & 0.5698          & \textbf{0.5538} & \textbf{0.5538} & 0.4553          & \textbf{0.8001} & \textbf{0.8476} & \textbf{0.6187} & \textbf{0.5671} \\ \hline
\end{tabular}
\caption{Given-$K$ methods clustering performance comparison results, where the best results are highlighted in bold and OOM denotes out of memory error.}
\label{tab:clustering_performance_givenk}
\end{table*}

\begin{table*}[t]
\centering
\setlength{\tabcolsep}{1mm}
\begin{tabular}{c|c|cccc}
\hline
Datasets    & SCMax  & Remove CL - I & Remove CL - II & Select Random Noise & Choose $\mathbf{G'}$ as result \\ \hline
MSRCv1      & \textbf{0.6238} & 0.6238             & 0.6238             & 0.1429              & 0.4952                  \\
MNIST    & \textbf{0.6004} & 0.3158             & 0.3902             & 0.6004              & 0.5596                  \\
Fashion   & \textbf{0.7542} & 0.1980             & 0.3913             & 0.7542              & 0.7194                  \\
YTF20       & \textbf{0.6493} & 0.2071             & 0.6493             & 0.3421              & 0.6455                  \\ \hline
\end{tabular}
\caption{The ablation experiment results on four representative datasets are presented, only showing the ACC metric results. The best result is displayed in bold. The complete results can be found in the Appendix.}
\label{tab:ablation_acc}
\end{table*}

\section{Experiments}

\subsection{Experimental Setup}

Regarding the datasets, we employ eight single-view datasets: MSRCv1\footnote{\url{https://www.microsoft.com/en-us/research/project/image-understanding}}, HW2\footnote{\url{https://cs.nyu.edu/roweis/data.html}}, Wiki\footnote{\url{http://www.svcl.ucsd.edu/projects/crossmodal/}}, MNIST\footnote{\url{http://yann.lecun.com/exdb/mnist/}}, Cifar10\footnote{\url{http://www.cs.toronto.edu/~kriz/cifar.html}}, Cifar100\footnote{\url{http://www.cs.toronto.edu/~kriz/cifar.html}}, Fashion \cite{xiaofashion}, YTF20\footnote{\url{https://www.cs.tau.ac.il/~wolf/ytfaces/}}, as shown in Table \ref{tab:dataset_description}. Regarding the comparison methods, since the proposed algorithm addresses the issue of $K$-value selection in parameter-free clustering, we select nine Non-$K$ clustering methods: FINCH \cite{sarfraz2019efficient}, COMIC \cite{peng2019comic}, BP \cite{averbuch2020border}, DenMune \cite{abbas2021denmune}, DeepDPM \cite{ronen2022deepdpm}, MPAASL \cite{dai2024multi}, TC \cite{yang2025autonomous}, Gauging-$\delta$ \cite{yao2025gauging} and AFCL \cite{zhang2025asynchronous}. Moreover, we also selected two Given-$K$ methods for reference comparison: DCGL \cite{chen2024deep} and DMAC \cite{wang2025towards}. Finally, we adopt four metrics to evaluate clustering quality: Accuracy (ACC), Normalized Mutual Information (NMI), Purity (PUR), and F-score. More information about the datasets, comparison methods, and implementation details is provided in the Appendix.

\subsection{Clustering Performance Comparison}

As shown in Table \ref{tab:clustering_performance}, \ref{tab:clustering_performance_givenk}, SCMax demonstrates superior clustering performance. Compared with Non-$K$ clustering methods, it achieves the highest scores in both ACC and F-score across all datasets, indicating its strong capability to uncover the true underlying cluster structures. For the NMI metric, SCMax also exhibits robust performance, significantly outperforming most competitors. Although it does not always achieve the best NMI on a few datasets such as MSRCv1, its results remain very close to the top-performing method, confirming its adaptability and effectiveness across diverse application scenarios. Meanwhile, for the PUR metric—which often favors over-segmentation—SCMax still delivers competitive results, reflecting its balanced performance across multiple evaluation perspectives. Compared with Given-$K$ clustering methods, it is important to note that such a comparison is inherently unfair: Given-$K$ methods operate with the prior knowledge of the cluster number $K$, whereas Non-$K$ methods must simultaneously perform cluster number estimation and clustering optimization, incurring additional $K$-value selection overhead. Even under this unfair comparison, our method still achieves highly competitive results and even attains the best performance on several datasets, including Wiki, Cifar10, Cifar100, Fashion, and YTF20. Overall, these results strongly validate the effectiveness and robustness of SCMax in practical tasks.

Due to the page limitation, the comparisons of estimated cluster number, running time and computational complexity for all methods are provided in the Appendix.

\subsection{Ablation Study}

To evaluate the effectiveness and design rationale of each component in SCMax, we conducted ablation studies. Table \ref{tab:ablation_acc} reports the ACC scores on four representative datasets, complete results are provided in the Appendix. We designed four ablation settings:

\begin{itemize}
    \item Remove CL-I: Removing the contrastive learning constraint Part I, retaining only the perturbation method based on pushing apart negative (different-class) samples.
    \item Remove CL-II: Removing the contrastive learning constraint Part II, retaining only the perturbation method based on pulling together positive (same-class) samples.
    \item Select Random Noise: Replacing perturbation with random noise sampled from $[-1,1]$, whose standard deviation matches that of the fixed representation $\mathbf{Z}$.
    \item Choose $\mathbf{G'}$ as Result: Selecting the cluster structure $\mathbf{G'}$ at the moment of consensus maximization as the final result.   
\end{itemize}

Experimental results show that removing or replacing any component leads to performance degradation, validating the necessity of each design. Notably, the comparison between "Remove CL-I" and "Remove CL-II" indicates that perturbations based on pushing apart negative samples have a more pronounced impact, suggesting that negative pair repulsion plays a more critical role in generating effective perturbations. The "Select Random Noise" setting leads to unstable and highly variable results, revealing the limitations of perturbations without structural constraints. Moreover, the "Choose $\mathbf{G'}$ as Result" setting consistently yields inferior performance compared to choosing $\mathbf{G}$, further indicating that the perturbed candidate cluster structures are less likely to produce optimal clustering results.

\subsection{Analysis}

In this subsection, we analyze SCMax from two perspectives: \textbf{the NNC score} and \textbf{loss convergence}. The NNC score effectively evaluates the latent cluster structure, reaching its maximum exactly at the closest ground-truth number of clusters, while SCMax exhibits stable and efficient training, with both the autoencoder and contrastive losses showing steady convergence. For details, see the Appendix.

\section{Conclusion}

The proposed SCMax framework addresses the true parameter-free clustering problem. By integrating hierarchical clustering, self-supervised representation learning, and nearest-neighbor-based consensus evaluation within a unified process, SCMax not only dynamically guides the generation of clustering number but also automatically evaluates the optimal clustering structure. Extensive experimental results demonstrate the superior performance of SCMax across multiple datasets. In future work, we aim to explore more advanced cluster number generation strategies to further enhance the accuracy and robustness of the resulting clustering structures. Furthermore, we plan to eliminate the reliance of the autoencoder-based representation learning module on predefined architectural structures.

\appendix

\twocolumn[
\begin{center}
    {\LARGE \textbf{Appendix}}
\end{center}
\vspace{1em}
]

\begin{table*}[!ht]
\centering
\begin{tabular}{c|cccccccc}
\hline
Methods  & MSRCv1     & HW2         & Wiki        & MNIST       & Cifar10     & Cifar100     & Fashion     & YTF20       \\ \hline
$K_{true}$  & 7          & 10          & 10          & 10          & 10          & 100          & 10          & 20          \\ \hline
FINCH    & 2          & 2           & 13          & 2           & 8           & 60           & 3           & 2           \\
COMIC    & 3          & 6           & 1786        & 20          & OOM         & OOM          & OOM         & OOM         \\
BP       & 3          & 2           & 15          & 6           & OOM         & OOM          & OOM         & OOM         \\
DenMune  & 19         & 138         & 247         & 450         & 5224        & 4590         & 6206        & 7491        \\
DeepDPM  & 1          & 1           & 1           & 1           & 11          & 31           & 33          & 86          \\
MPAASL   & 19         & \textbf{11} & 19          & 19          & OOM         & OOM          & OOM         & OOM         \\
TC       & 37         & 24          & 21          & 26          & OOM         & OOM          & OOM         & OOM         \\
Gauging-$\delta$ & 4          & 1           & 1           & 1           & OOM         & OOM          & OOM         & OOM         \\
AFCL     & \textbf{7} & 18          & 10          & \textbf{10} & OOM         & OOM          & OOM         & OOM         \\
DCGL     & —          & —           & —           & —           & —           & —            & —           & —           \\
DMAC     & —          & —           & —           & —           & —           & —            & —           & —           \\ \hline
SCMax    & 8          & 8           & \textbf{10} & 15          & \textbf{11} & \textbf{106} & \textbf{13} & \textbf{25} \\ \hline
\end{tabular}
\caption{Estimated number of clusters by each method, where the result closest to the ground truth is highlighted in bold.}
\label{tab:estimated_number}
\end{table*}

\begin{table*}[!ht]
\centering
\begin{tabular}{ccccccccc}
\hline
Methods  & MSRCv1        & HW2           & Wiki          & MNIST         & Cifar10         & Cifar100        & Fashion         & YTF20           \\ \hline
FINCH    & \textbf{0.03} & \textbf{0.11} & \textbf{0.12} & \textbf{0.48} & \textbf{23.52}  & \textbf{22.28}  & \textbf{30.73}  & \textbf{37.18}  \\
COMIC    & 2.71          & 24.30         & 32.61         & 166.69        & OOM             & OOM             & OOM             & OOM             \\
BP       & \textit{0.10} & 2.50          & \textit{4.41} & 13.12         & OOM             & OOM             & OOM             & OOM             \\
DenMune  & 0.62          & 3.78          & 5.34          & 10.04         & \textit{141.44} & \textit{132.72} & \textit{183.61} & \textit{173.84} \\
DeepDPM  & 629.19        & 656.86        & 682.84        & 745.03        & 4044.09         & 6554.37         & 8269.58         & 12128.43        \\
MPAASL   & 257.05        & 3511.38       & 6340.23       & 16718.54      & OOM             & OOM             & OOM             & OOM             \\
TC       & 1.15          & \textit{0.84} & 21.50         & \textit{3.66} & OOM             & OOM             & OOM             & OOM             \\
Gauging-$\delta$ & 5.73          & 3311.67       & 3573.84       & 38974.31      & OOM             & OOM             & OOM             & OOM             \\
AFCL     & 539.85        & 2925.35       & 293.47        & 9276.38       & OOM             & OOM             & OOM             & OOM             \\
DCGL     & 7.49          & 303.30        & 742.11        & 3209.28       & OOM             & OOM             & OOM             & OOM             \\
DMAC     & 7.59          & 18.87         & 26.22         & 48.82         & OOM             & OOM             & OOM             & OOM             \\
SCMax    & 4.94          & 16.86         & 14.46         & 43.41         & 1168.78         & 1163.23         & 1821.79         & 1052.95         \\ \hline
\end{tabular}
\caption{Running time of each method, where the shortest running time is highlighted in bold and the second shortest is shown in italics.}
\label{tab:running_time}
\end{table*}

\section{Details of Employed Datasets}

This section provides detailed descriptions of the eight datasets used in our experiments.

\begin{itemize}
    \item MSRCv1: An image segmentation and object recognition dataset containing multiple semantic object categories.
    \item HW2: A handwritten digit image dataset containing multiple digit categories.
    \item Wiki: A multimodal dataset containing multiple semantic topic categories.
    \item MNIST: A heterogeneous multi-view handwritten digit dataset containing multiple digit categories.
    \item Cifar10: A natural image dataset containing multiple common object categories.
    \item Cifar100: A fine-grained version of the Cifar10 dataset containing multiple specific object subcategories. 
    \item Fashion: A fashion image dataset containing multiple clothing categories.
    \item YTF20: A face image dataset containing multiple identity categories.
\end{itemize}

\section{Details of Comparison Methods}

This section provides detailed descriptions of the eleven comparison methods employed in our experiments. 

\begin{itemize}
    \item FINCH: A Non-$K$ hierarchical clustering algorithm based on nearest neighbors, capable of generating cluster partitions at different levels. Since this method does not provide an evaluation mechanism for partition selection, we choose the clustering number corresponding to the last partition as the final result.
    \item COMIC: A Non-$K$ multi-view clustering method that integrates geometric consistency and clustering assignment consistency, achieving efficient clustering without predefining the number of clusters. Moreover, we follow the original paper’s setting and use the average length of the shortest 90\% similarity graph edges as the threshold for final cluster structure formation.
    \item BP: A Non-$K$ clustering method that explicitly reveals the core cluster structure by progressively peeling off boundary points. For fairness, we follow the original paper’s autoencoder configuration, reducing features to 10 dimensions and training for 100 epochs. Other parameter settings are in accordance with the open-source code of the original paper.
    \item DenMune: A Non-$K$ clustering method based on neighbor density consistency. The paper indicates the method is insensitive to hyperparameters, thus, we set the number of $k$-nearest neighbors to 4 during reproduction, consistent with the original paper’s demo dataset setting.
    \item DeepDPM: A deep Non-$K$ clustering method based on a split-merge strategy that can automatically infer the number of clusters during training. Following the settings of the original paper, we use the same autoencoder architecture as SCMax, train for 200 epochs, reduce the feature dimensions to 128, and set the initial number of clusters to the default value of 1.
    \item MPAASL: A Non-$K$ multi-view clustering method combining graph consistency modeling with reinforcement learning mechanisms. We obtained the source code by contacting the authors and set the view voting mechanism to "all" during reproduction.
    \item TC: A Non-$K$ clustering method based on an automatic merge strategy guided by inter-cluster mass and distance, which can adaptively discover the cluster structure without predefining the cluster number. All parameter settings are in accordance with the open-source code of the original paper.
    \item Gauging-$\delta$: A Non-$K$ clustering method based on an adaptive merge strategy. The method introduces a gauging function that leverages statistical and environmental information of clusters to adaptively determine whether two clusters can be merged. Following the original paper, We employed a simulated CNN architecture to reduce the features to 128 dimensions.
    \item AFCL: A Non-$K$ federated clustering method designed for asynchronous and heterogeneous client scenarios. The method introduces redundant seed points as communication media and gradually prunes them to reveal the global cluster distribution without requiring the true number of clusters. In reproduction, we followed the original settings: we removed the missing samples and normalized all the data.
    \item DMAC: A Given-$K$ multi-view clustering method based on learnable anchors. Unlike traditional approaches with fixed anchors, DMAC employs noise-driven anchor learning and anchor graph convolution, ensuring anchors adapt during training.
    \item DCGL: A Given-$K$ clustering method combining autoencoder and graph convolution with dual contrastive guidance. The method enhances discriminability by integrating feature-level contrast and cluster-level contrast, jointly optimizing both original features and learned graph structures.
\end{itemize}

For SCMax, the architecture of the autoencoder is: dim(X)-500-500-2000-256. The autoencoder is trained for 200 epochs, with 50 epochs for contrastive learning constraints in each round. We set the batch size to 256, learning rate to 0.0003, and the random seed to 3407 \cite{picard2021torch}. Moreover, all experiments are conducted on a machine configured with 60 GB of RAM, a 16 vCPU AMD EPYC 9654 processor (96 cores), and a virtual GPU with 32 GB of VRAM\footnote{\url{https://www.autodl.com/home}}.

\begin{table*}[!ht]
\centering
\begin{tabular}{cccc}
\hline
Method    & Memory Cost      & Time Complexity     & Max Reported \\ \hline
FINCH     & O($N$)           & O($N \log N$)       & 63896        \\
COMIC     & O($N^2$)         & O($N^2$)            & 5000         \\
BP        & O($ND$)          & O($N^2$)            & 5000         \\
DenMune   & O($NK$)          & O($N^2K$)           & 63896        \\
DeepDPM   & O($NK$)          & O($NKD^2$)          & 63896        \\
MPAASL    & O($N^2$)         & O($N^3$)            & 5000         \\
TC        & O($N^2$)         & O($N \log N$)       & 5000         \\
Gauging-$\delta$ & O($N^2$)  & O($N^2 \log N$)     & 5000         \\
AFCL      & O($ND$)          & O($KNDP$)           & 5000         \\
DMAC      & O($ND$)          & O($NM$)             & 5000         \\
DCGL      & O($N^2$)         & O($N^3$)            & 5000         \\ \hline
SCMax     & O($K + 2K^2 + B^2$) & O($K \log K + K + B^2 + K^3$) & 63896        \\ \hline
\end{tabular}
\caption{The computational complexity of each method, where Max Reported denotes the maximum dataset size that can be processed, $N$ represents the number of samples, $M$ denotes the number of anchors, $K$ is the number of clusters, $B$ is the batch size, $D$ is the feature dimension, and $P$ is the number of clients.}
\label{tab:computational}
\end{table*}

\begin{table*}[!ht]
\centering
\begin{tabular}{c|c|cccc}
\hline
Datasets    & SCMax  & Remove CL - I & Remove CL - II & Select Random Noise & Choose $\mathbf{G'}$ as result \\ \hline
\multicolumn{6}{c}{ACC}                                                                                        \\ \hline
MSRCv1      & \textbf{0.6238} & 0.6238             & 0.6238             & 0.1429              & 0.4952                  \\
HW2  & 0.6115 & 0.5115             & 0.2955             & 0.2955              & \textbf{0.6465}                  \\
Wiki        & \textbf{0.5976} & 0.3814             & 0.5967             & 0.5967              & 0.5963                  \\
MNIST & \textbf{0.6004} & 0.3158             & 0.3902             & 0.6004              & 0.5596                  \\
Cifar10    & \textbf{0.8703} & 0.2458             & 0.8703             & 0.8703              & 0.8261                  \\
Cifar100   & \textbf{0.9061} & 0.9061             & 0.9061             & 0.9061              & 0.8480                  \\
Fashion      & \textbf{0.7542} & 0.1980             & 0.3913             & 0.7542              & 0.7194                  \\
YTF20       & \textbf{0.6493} & 0.2071             & 0.6493             & 0.3421              & 0.6455                  \\ \hline
\multicolumn{6}{c}{NMI}                                                                                        \\ \hline
MSRCv1      & \textbf{0.5597} & 0.5597             & 0.5597             & 0.0000              & 0.4881                  \\
HW2  & 0.6436 & 0.6574             & 0.5329             & 0.5329              & \textbf{0.6992}                  \\
Wiki        & \textbf{0.5609} & 0.5100             & 0.5609             & 0.5609              & 0.5528                  \\
MNIST & 0.7021 & 0.6436             & 0.5703             & 0.7021              & \textbf{0.7059}                  \\
Cifar10    & \textbf{0.7714} & 0.5507             & 0.7714             & 0.7714              & 0.7489                  \\
Cifar100   & \textbf{0.9801} & 0.9801             & 0.9801             & 0.9801              & 0.9690                  \\
Fashion      & \textbf{0.7479} & 0.5352             & 0.6084             & 0.7479              & 0.7131                  \\
YTF20       & \textbf{0.8605} & 0.7031             & 0.8065             & 0.5535              & 0.7982                  \\ \hline
\multicolumn{6}{c}{PUR}                                                                                        \\ \hline
MSRCv1      & \textbf{0.6476} & 0.6476             & 0.6476             & 0.1429              & 0.6000                  \\
HW2  & 0.6115 & \textbf{0.8065}             & 0.2955             & 0.2955              & 0.6465                  \\
Wiki        & 0.6403 & \textbf{0.6647}             & 0.6403             & 0.6403              & 0.6270                  \\
MNIST & 0.7658 & \textbf{0.8970}             & 0.3902             & 0.7658              & 0.8370                  \\
Cifar10    & 0.8922 & \textbf{0.8961}             & 0.8922             & 0.8922              & 0.8824                  \\
Cifar100   & \textbf{0.9318} & 0.9318             & 0.9318             & 0.9318              & 0.9287                  \\
Fashion      & 0.8497 & \textbf{0.8696}             & 0.3913             & 0.8497              & 0.8355                  \\
YTF20       & 0.8071 & \textbf{0.9842}             & 0.8071             & 0.3421              & 0.8269                  \\ \hline
\multicolumn{6}{c}{F-score}                                                                           \\ \hline
MSRCv1      & \textbf{0.5698} & 0.5698             & 0.5698             & 0.0357              & 0.3821                  \\
HW2  & \textbf{0.5538} & 0.2588             & 0.1377             & 0.1377              & 0.5473                  \\
Wiki        & \textbf{0.5538} & 0.1641             & 0.5538             & 0.5538              & 0.5424                  \\
MNIST & \textbf{0.4553} & 0.0901             & 0.2580             & 0.4553              & 0.3552                  \\
Cifar10    & \textbf{0.8001} & 0.0256             & 0.8001             & 0.8001              & 0.7170                  \\
Cifar100   & \textbf{0.8476} & 0.8476             & 0.8476             & 0.8476              & 0.6646                  \\
Fashion      & \textbf{0.6178} & 0.0168             & 0.2451             & 0.6178              & 0.5465                  \\
YTF20       & \textbf{0.5671} & 0.0267             & 0.5671             & 0.1671              & 0.5368                  \\ \hline
\end{tabular}
\caption{The complete ablation experiment results, including four evaluation indicators: ACC, NMI, PUR, and F-score. The best result is displayed in bold.}
\label{tab:ablation_full}
\end{table*}

\begin{table*}[!ht]
\centering
\begin{tabular}{c|c|c|cccc}
\hline
Datasets    & $K_{true}$ & SCMax & Remove CL - I & Remove CL - II & Select Random Noise & Choose $\mathbf{G'}$ as result \\ \hline
MSRCv1      & 7       & \textbf{8}     & 8                  & 8                  & 1                   & 10                      \\
HW2  & 10      & \textbf{8}     & 24                 & 3                  & 3                   & 7                          \\
Wiki        & 10      & \textbf{10}    & 28                 & 10                 & 10                  & 9                      \\
MNIST & 10      & \textbf{15}    & 51                 & 4                  & 15                  & 19                      \\
Cifar10    & 10      & \textbf{11}    & 148                & 11                 & 11                  & 12                       \\
Cifar100   & 100     & \textbf{106}   & 106                & 106                & 106                 & 129                     \\
Fashion      & 10      & \textbf{13}    & 190                & 4                  & 13                  & 14                      \\
YTF20       & 20      & \textbf{25}    & 262                & 25                 & 6                   & 27                      \\ \hline
\end{tabular}
\caption{The ablation experiment results of the number of clusters predicted by all methods. The best result is displayed in bold.}
\label{tab:ablation_number}
\end{table*}

\begin{table*}[!ht]
\centering
\begin{tabular}{c|c|cccc}
\hline
Datasets    & SCMax   & Remove CL - I & Remove CL - II & Select Random Noise & Choose $\mathbf{G'}$ as result \\ \hline
MSRCv1      & 4.94    & 5.06               & 4.96               & \textbf{4.48}                & 4.94                    \\
HW2  & 16.86   & 14.84              & 17.99              & \textbf{9.87}                & 16.86                   \\
Wiki        & 14.46   & 17.59              & 17.43              & \textbf{10.31}               & 14.46                   \\
MNIST & 43.41   & 38.25              & 42.51              & \textbf{21.33}               & 43.41                   \\
Cifar10    & 1168.78 & 1128.83            & 1202.92            & \textbf{955.88}              & 1168.78                 \\
Cifar100   & 1163.23 & 1122.84            & 1165.80            & \textbf{1014.40}             & 1163.23                 \\
Fashion      & 1821.79 & 1720.86            & 1817.21            & \textbf{1508.49}             & 1821.79                 \\
YTF20       & 1052.95 & 1049.76            & 1104.84            & \textbf{816.60}              & 1052.95                 \\ \hline
\end{tabular}
\caption{The ablation experiment results of the running times for all methods. The best result is displayed in bold.}
\label{tab:ablation_time}
\end{table*}

\begin{figure*}[!ht]
\begin{center}
{
\centering
\subfloat[MSRCv1]{{\includegraphics[width=0.48\columnwidth]{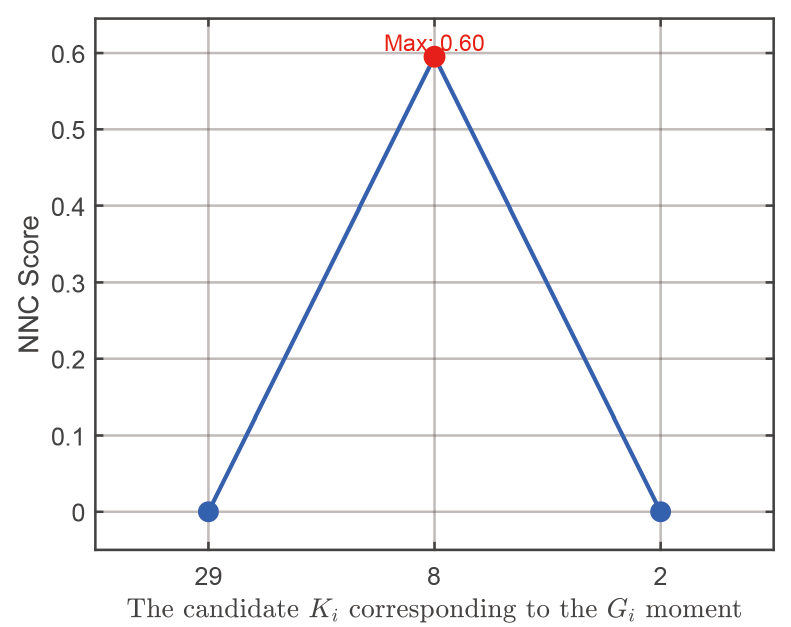}}}
\subfloat[HW2]{{\includegraphics[width=0.48\columnwidth]{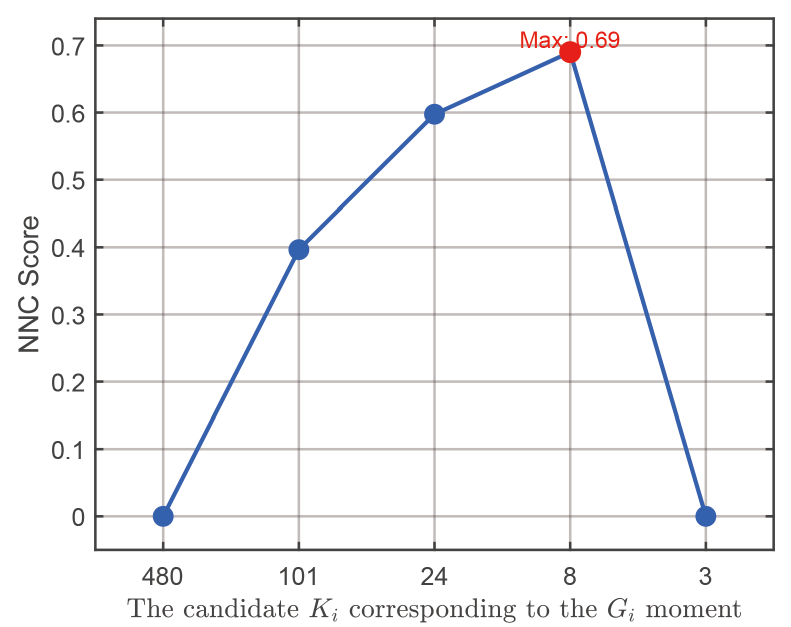}}}
\subfloat[Wiki]{{\includegraphics[width=0.48\columnwidth]{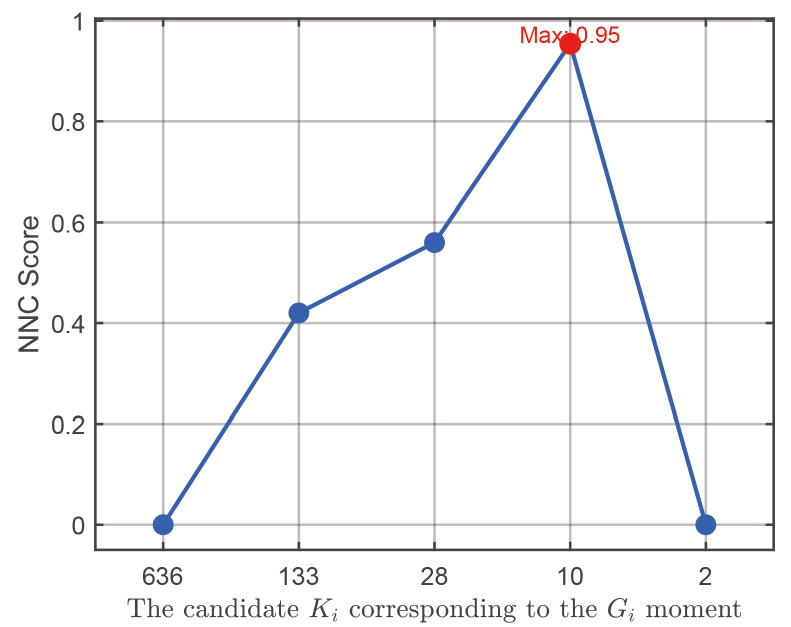}}}
\subfloat[MNIST]{{\includegraphics[width=0.48\columnwidth]{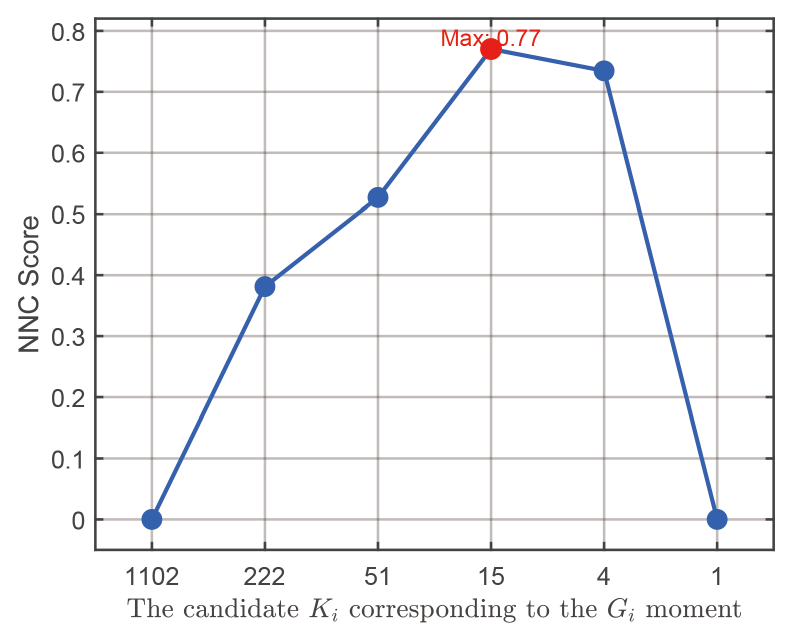}}}

\subfloat[Cifar10]{{\includegraphics[width=0.48\columnwidth]{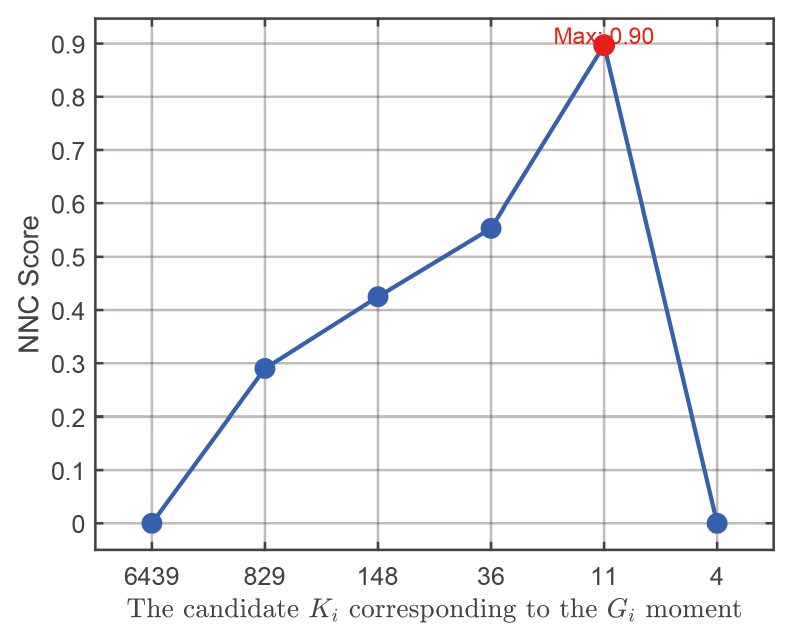}}}
\subfloat[Cifar100]{{\includegraphics[width=0.48\columnwidth]{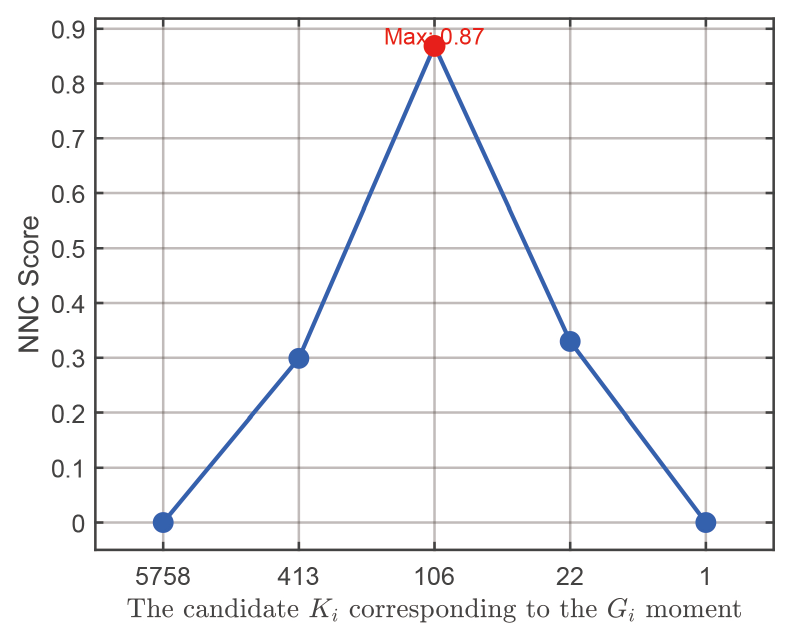}}}
\subfloat[Fashion]{{\includegraphics[width=0.48\columnwidth]{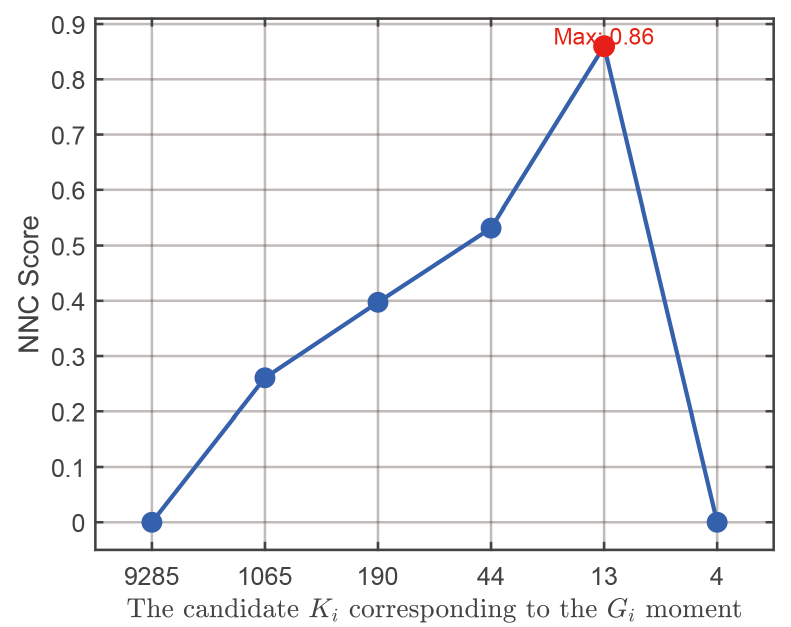}}}
\subfloat[YTF20]{{\includegraphics[width=0.48\columnwidth]{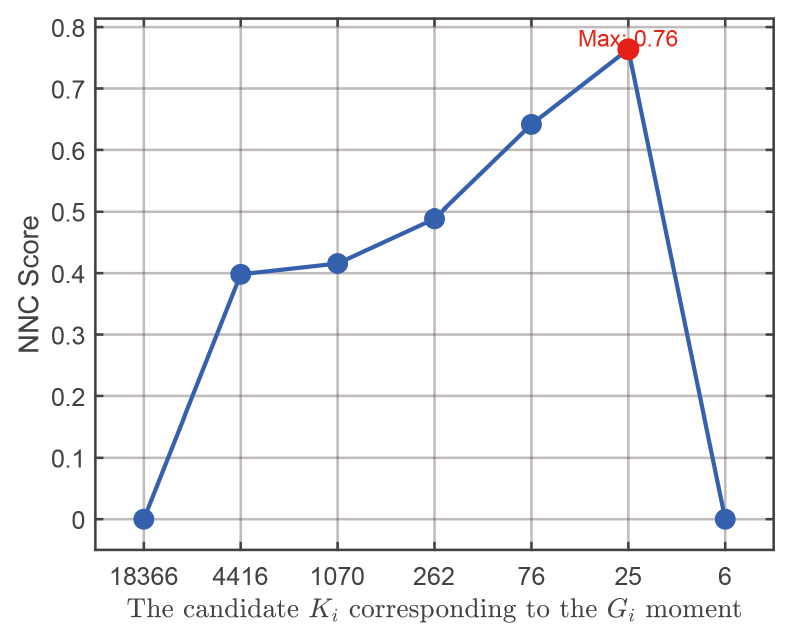}}}
\caption{The NNC scores of the candidate $K_i$ values corresponding to the $G_i$ time on all datasets. }
\label{fig:score_all}
}
\end{center}
\end{figure*}

\section{Additional Comparisons of Clustering Performance}

In this subsection, we present additional result analyses of our algorithm compared with eleven comparison methods across eight datasets. The supplementary analysis includes the estimated number of clusters, the running time and the computational complexity.

\textbf{Estimated Cluster Number Comparison.} As shown in Table \ref{tab:estimated_number}, SCMax achieves the most accurate estimation of the true number of clusters on most datasets, demonstrating its superior capability in cluster number inference. Notably, on large-scale datasets with over ten thousand samples, SCMax significantly outperforms other methods. This advantage stems from its parameter-free and efficient unified framework, enabling it to better adapt to complex high-dimensional data. By contrast, other methods show various limitations: FINCH and BP lack effective evaluation mechanisms and tend to produce trivial solutions, grouping most samples into a single cluster. COMIC, DenMune, and TC are overly dependent on sensitive explicit or implicit threshold parameters, which, if misconfigured, lead to local optima and oversegmentation. DeepDPM requires an initial number of clusters; inaccurate initialization can misguide the entire clustering process, especially on small datasets. MPAASL restricts the search space to a predefined range of $K$, limiting its generalizability beyond the specified range.

\textbf{Running Time Comparison.} As shown in Table \ref{tab:running_time}, SCMax not only estimates cluster numbers accurately but also delivers outstanding computational efficiency. On all datasets, its runtime remains highly competitive, particularly on large-scale datasets, where it completes clustering in a short time, showcasing strong scalability. In contrast, comparison methods present various computational challenges: DeepDPM incurs high training time due to its complex deep modeling. COMIC, BP, MPAASL, TC, DCFL, and DMAC frequently experience out-of-memory (OOM) issues due to high computational demands. FINCH often shows extremely low runtime because it simply selects the last candidate $K$ as the final result due to its lack of an effective evaluation mechanism. DenMune frequently stops early due to its sensitivity to improper threshold settings, leading to oversegmentation and relatively low computational cost.

\textbf{Computational Complexity Comparison.} As shown in Table \ref{tab:computational}, SCMax achieves highly competitive computational complexity while maintaining optimal clustering performance. First, compared with Non-$K$ clustering methods, the complexity of SCMax mainly depends on the number of clusters $K$ and the batch size $B$, rather than the total number of samples $N$ or the feature dimension $D$. This design enables SCMax to scale effectively to large datasets. In contrast, most Non-$K$ methods (such as COMIC, BP, MPAASL, TC, Gauging-$\delta$, and AFCL) exhibit quadratic or even cubic growth in complexity as $N$ or $D$ increases, leading to excessive memory consumption and frequent out-of-memory errors. Second, when compared with Given-$K$ clustering methods, this comparison is inherently unfair: Given-$K$ methods operate with the prior knowledge of the cluster number $K$, whereas Non-$K$ methods must simultaneously perform cluster number estimation and clustering optimization, incurring additional $K$-value selection overhead. Even under such unfair conditions, SCMax still demonstrates lower overall complexity than Given-$K$ methods such as DCGL and DMAC. Notably, DCGL and DMAC can only run successfully on datasets with up to 5,000 samples, while they fail to operate on larger datasets containing tens of thousands of samples. In summary, SCMax maintains superior clustering performance while achieving remarkable computational efficiency and scalability.

\section{Ablation Study on All Datasets}

To further validate the stability and effectiveness of each component in the SCMax framework across different data environments, we supplement the appendix with four sets of ablation experiments conducted on all eight datasets. As shown in Tables \ref{tab:ablation_full}, \ref{tab:ablation_number}, and \ref{tab:ablation_time}, all ablation settings exhibit a performance decline trend across different datasets, confirming the indispensable roles of each module in the overall framework. Besides the main results, some interesting observations can be made: First, although “Select Random Noise” as a perturbation method shows very unstable performance, occasionally good results inspire us to simplify the perturbation strategy in future designs, indicating potential computational efficiency gains. Second, the results from “Choose $\mathbf{G'}$ as result” demonstrate that the cluster number evaluation can only accurately identify the best candidate $K$ but cannot evaluate cluster numbers outside the candidate set. Therefore, clustering performance depends solely on the quality of candidate $K$s generated by the cluster number generation module, which guides future explorations towards higher-quality cluster number generation strategies.

\section{NNC Score Analysis}

In this section, we visualize the NNC scores corresponding to the candidate cluster structures $\mathbf{G}_i$ at different moments and investigate the reliability of the nearest-neighbor consensus evaluation. As illustrated in Figure \ref{fig:score_all}, the variation of the NNC score shows that as $K_i$ approaches the true number of clusters, the NNC score increases steadily, and it decreases as the estimated number moves further away from the ground truth. This verifies the effectiveness of nearest-neighbor consensus in evaluating cluster structure. Remarkably, the Maximum of the NNC score aligns precisely with the closest ground-truth cluster structure, indicating that the metric can accurately identify the optimal number of clusters without any hyperparameter settings. These results show our metric can adaptively and reliably estimate data’s intrinsic structure without any parameters.

\section{Convergence Analysis}

\begin{figure}[!ht]
\begin{center}
{
\centering
\subfloat[AE loss $L_R$ on Cifar10]{{\includegraphics[width=0.48\columnwidth]{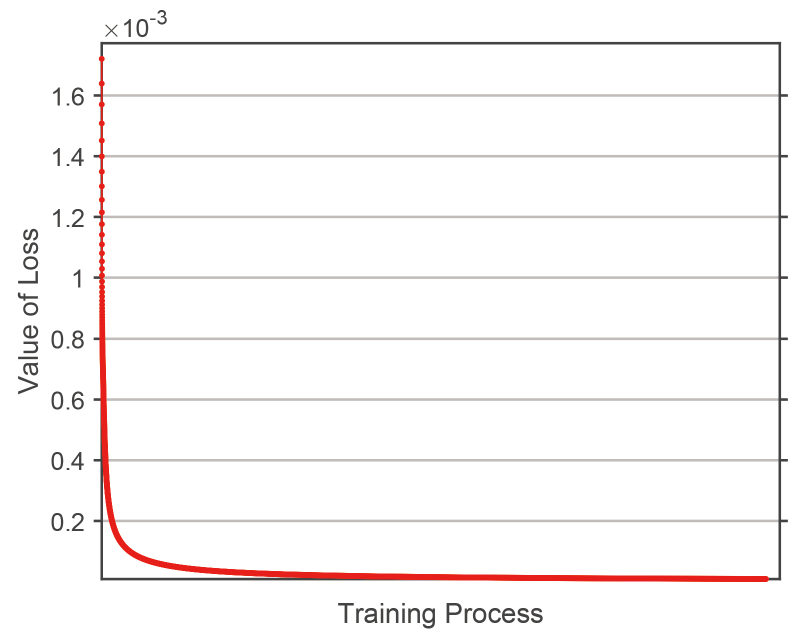}}}
\subfloat[AE loss $L_R$ on Cifar100]{{\includegraphics[width=0.48\columnwidth]{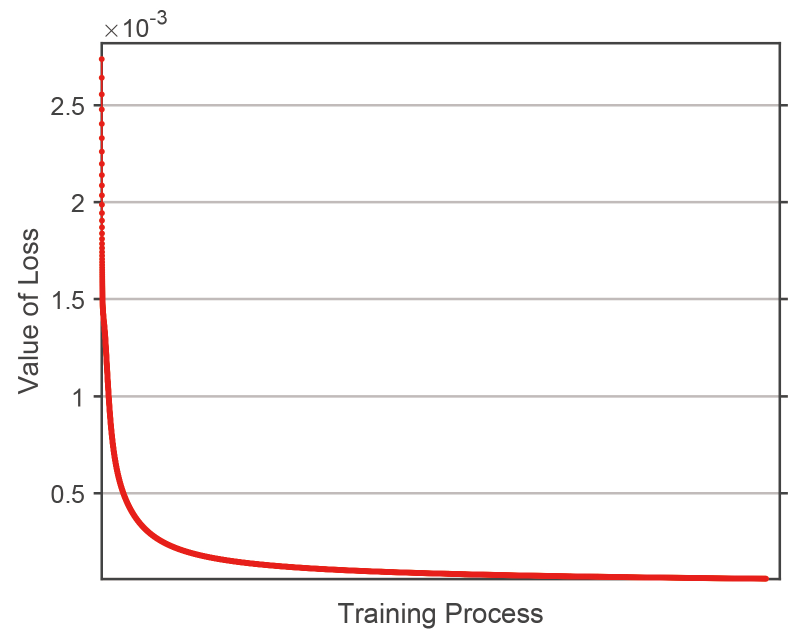}}}

\subfloat[CL loss $L_Q$ on Cifar10]{{\includegraphics[width=0.48\columnwidth]{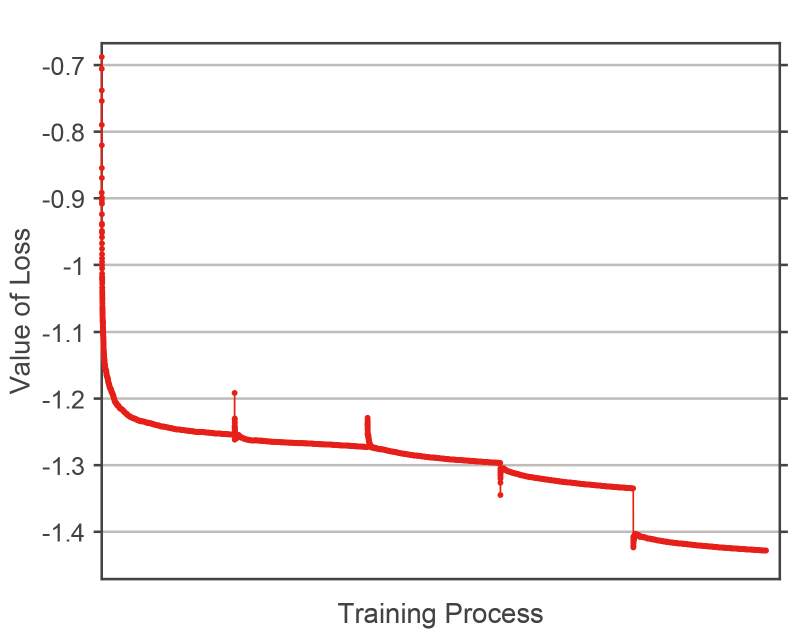}}}
\subfloat[CL loss $L_Q$ on Cifar100]{{\includegraphics[width=0.48\columnwidth]{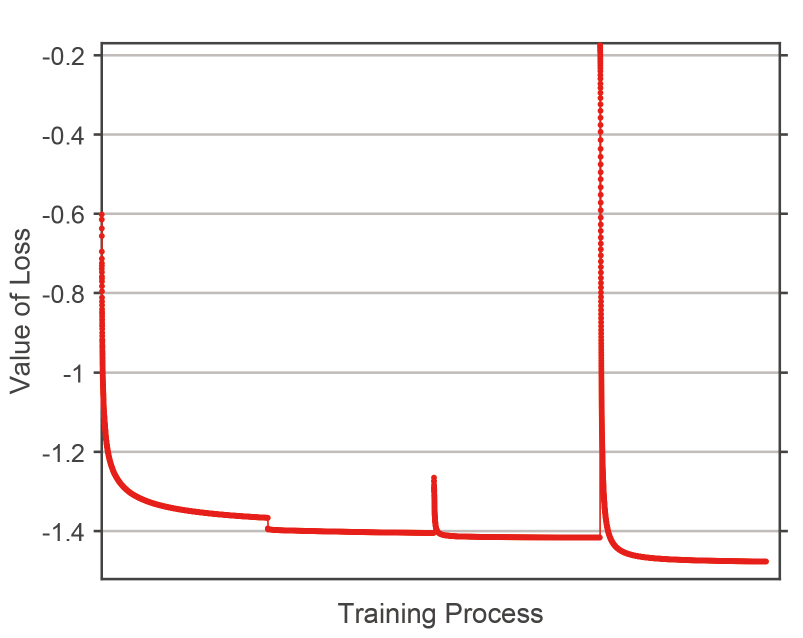}}}
\caption{Convergence curves on the Cifar10 and Cifar100 datasets. Here, each iteration represents one time of network training, including both the autoencoder and contrastive learning constraints.}
\label{fig:loss}
}
\end{center}
\end{figure}

To analyze the convergence of SCMax, we track the loss values during training on Cifar10 and Cifar100 as the number of iterations increases. As shown in Figure \ref{fig:loss}, the proposed SCMax exhibits a clear convergence trend. The total loss consists of two parts: the loss $L_R$ from the autoencoder (AE) and the loss $L_Q$ from contrastive learning (CL) constraints. Both losses consistently decrease throughout training, indicating continuous improvement in dimensionality reduction and feature optimization. Specifically, the autoencoder loss drops rapidly in the early stages, demonstrating that the model quickly captures the basic structure of the data. Meanwhile, the contrastive learning loss also decreases and stabilizes across different candidate $K$ values, showing the model's persistent ability to enhance feature discriminability. Overall, SCMax demonstrates a smooth and efficient training process across both datasets, further validating the soundness and effectiveness of its training strategy.

\section*{Acknowledgements}

This work is supported by the National Science Fund for Distinguished Young Scholars of China (No. 62325604), and the National Natural Science Foundation of China (No. 62276271 and 62376039).

\bibliography{zhang}

\end{document}